\documentclass[11pt]{article}
\usepackage[T1]{fontenc}
\usepackage[utf8]{inputenc}
\usepackage{lmodern,amsmath,amssymb,booktabs,array,graphicx}
\usepackage[margin=1in]{geometry}
\usepackage{microtype}
\usepackage{pgfplots}
\usetikzlibrary{arrows.meta}
\pgfplotsset{compat=1.18}
\usepackage[colorlinks=true,linkcolor=blue!50!black,citecolor=blue!50!black,urlcolor=blue!50!black]{hyperref}
\usepackage{placeins}
\title{Inherited Learning in an Artificial Ecology: \\ How Controls and Update Allocation Shape Benefits}
\author{Xuening Wu\thanks{Email: \href{mailto:xuening.wu@pfizer.com}{\texttt{xuening.wu@pfizer.com}}}\\Pfizer, Shanghai, China\and Lei Li\\Pfizer, Shanghai, China\and Shan Yu\\Independent Researcher}
\date{}
\begin{document}
\maketitle
\begin{abstract}
Learning can improve an individual's behavior, yet a population risks losing that experience whenever individuals die and are replaced. Inheriting learned preferences offers a way to preserve useful behavior across generations, raising a question for artificial populations: when does inheritance improve collective performance, and how can its benefits be measured fairly? The challenge is that inheritance changes not only offspring behavior but also survival, reproduction, and opportunities for further hereditary updates. Random controls with equal update magnitudes may therefore yield misleading comparisons if they alter different states or obey different stability constraints. We investigate this problem in a resource-limited artificial ecology, combining structured random controls with interventions on newborn preferences and the allocation of hereditary updates. Preserving the state structure of random updates substantially narrows the apparent inheritance advantage, while a conditional establishment-speed benefit remains. Preference erasure and faster-learning compensation support a contribution from reduced offspring relearning. Update allocation also changes the comparison: event quotas and common time cutoffs can reverse rankings, although they also change realized update amounts. With update count and cumulative magnitudes matched, staged release improves occupancy but does not achieve the prespecified establishment criterion. These findings provide a framework for distinguishing the value of inherited preferences from the effects of control design and update allocation, clarifying how inherited learning should be evaluated in artificial populations.
\end{abstract}

\section{Introduction}
How can a population retain useful experience when its members die and are replaced? Passing learned parameters to offspring can reduce the need to reconstruct behavior, but a better initial policy also changes survival, reproduction, and later opportunities for inheritance. Evaluating inherited learning therefore requires connecting individual behavior to population dynamics.

Learning-mediated evolution and direct transmission of acquired traits have a substantial history \cite{hinton1987,rivoire2014}. Evolutionary robotics studies have compared inherited controllers with learning from scratch and examined newborn performance \cite{jelisavcic2019,luo2023}. Our question is how to distinguish the benefit of learned update direction from the experimental choices used to measure it.

Three coupled issues motivate the study. Equal-norm random perturbations can redistribute change across states that an agent cannot encounter. Directed and random updates can also differ in stability, because proportional directed writeback moves toward learned values while random rotation need not obey the same bounds. Finally, reproduction determines when hereditary updates occur, so equal event quotas can stop new hereditary updates at different times in different populations. These issues complicate the interpretation of a directed--random performance gap.

We study a resource-limited artificial ecology with local learning, stochastic death, and reproduction on a two-dimensional torus. Individuals choose among resting, harvesting, and dividing. An inherited table $g$ initializes lifetime action values $\theta$, and reproduction can transmit part of $\theta-g$. External ecological rules remain fixed, while local resources and population composition evolve. This computational testbed enables interventions on inheritance without an externally imposed population-level fitness selection operator.

\begin{figure}[!htb]
\centering
\begin{tikzpicture}[
  box/.style={draw=blue!45!black,rounded corners=3pt,fill=blue!4,
    text width=3.95cm,minimum height=1.25cm,align=center,inner sep=5pt,font=\footnotesize},
  test/.style={box,draw=teal!60!black,fill=teal!4,minimum height=1.7cm,anchor=north},
  flow/.style={-{Latex[length=1.8mm]},thick,draw=blue!60!black}]
\node[box] (birth) at (0,0) {\textbf{Inherited initialization}\\[2pt]
Hereditary table $g$\\Newborn values $\theta\leftarrow g$};
\node[box] (learn) at (4.9,0) {\textbf{Lifetime learning}\\[2pt]
Rest / harvest / divide\\$\Delta\theta_{s,a}=\eta(r-\theta_{s,a})$};
\node[box] (child) at (9.8,0) {\textbf{Local reproduction}\\[2pt]
$d=\lambda(\theta-g)$\\$g_{\rm child}=g_{\rm parent}+d$};
\draw[flow] (birth.east)--(learn.west);
\draw[flow] (learn.east)--(child.west);
\draw[flow] (child.north)--++(0,.38)-|node[pos=.25,above,font=\footnotesize]{Next generation} (birth.north);
\node[draw=gray!60,rounded corners,fill=gray!6,text width=13.75cm,
 align=center,font=\footnotesize,inner sep=5pt] (eco) at (4.9,-1.55)
 {\textbf{Shared ecology: 2D toroidal lattice}\\
 Resources, energy costs, death, and reproduction};
\draw[flow] (eco.north)--(learn.south);
\node[test] (geometry) at (0,-2.35)
 {\textbf{1. Update controls}\\[2pt]
Directed vs. random\\
State-preserving control\\
Shared stability bounds};
\node[test] (expression) at (4.9,-2.35)
 {\textbf{2. Newborn preferences}\\[2pt]
Keep vs. erase\\
Faster-learning rescue};
\node[test] (allocation) at (9.8,-2.35)
 {\textbf{3. Update allocation}\\[2pt]
Event vs. time cutoff\\
Early vs. staged updates};
\node[draw=gray!60,rounded corners,fill=gray!6,text width=13.75cm,
 align=center,font=\footnotesize,inner sep=5pt] (measure) at (4.9,-4.85)
 {\textbf{Outcomes and realized budgets}\\
Establishment, occupancy, extinction; births and hereditary change};
\draw[flow] (geometry.south)--(geometry.south |- measure.north);
\draw[flow] (expression.south)--(measure.north);
\draw[flow] (allocation.south)--(allocation.south |- measure.north);
\end{tikzpicture}
\caption{Study overview. Lifetime learning can enter offspring initialization through proportional writeback (top; $\mu=0$ in the central experiments). Controls preserve state-wise update structure and impose shared stability bounds. Newborn interventions test preference expression and faster-learning rescue. Allocation experiments compare event and time cutoffs, then early and staged release with matched counts and cumulative magnitudes. These interventions estimate distinct effects. Stopping writeback preserves lifetime learning and inheritance of existing $g$.}
\label{fig:overview}
\end{figure}
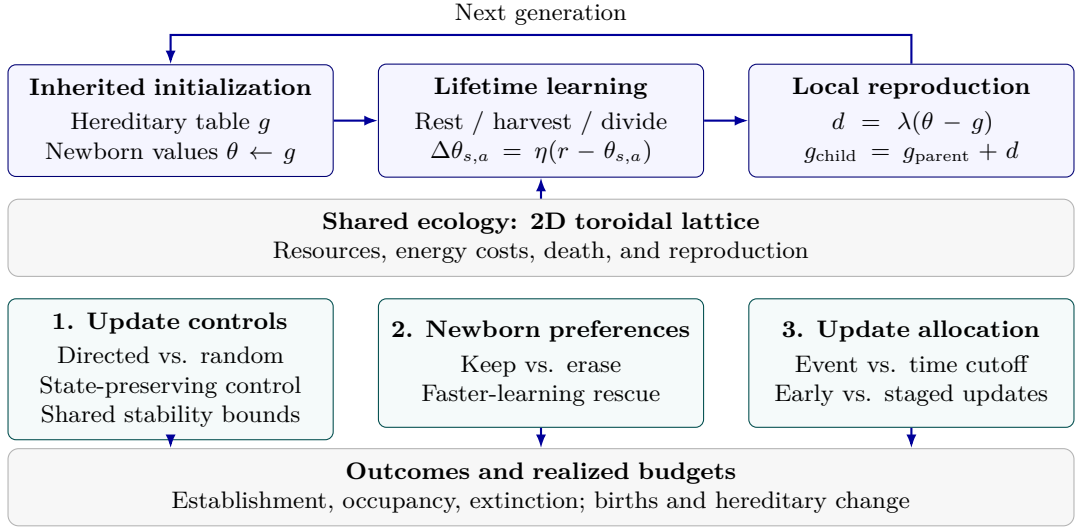

The evidence follows three questions (Figure~\ref{fig:overview}). First, does a benefit remain under a stronger random control? At a fixed 512-event budget, preserving each state's update structure sharply narrows the attainment gap relative to whole-table controls, while retaining a speed advantage. Second, does newborn expression contribute to that benefit? In a separate seed block, preference erasure impairs establishment and faster newborn learning compensates for much of the loss. Third, does the allocation of hereditary updates change the outcome? Proportional writeback retains continuous benefits under common stability constraints, but event quotas and shared time cutoffs can reverse the ranking. With update count and cumulative magnitudes matched, staged release improves occupancy without achieving the prespecified establishment threshold.

Our contribution is this sequence of controls and interventions, supported by quota, representation, and ecological sensitivity analyses. The comparisons estimate distinct effects: matching candidate geometry does not match diverging histories, and an occupancy improvement is not evidence of faster establishment. Together they bound the benefit of inherited preferences in this ecology and provide a framework for evaluating it in other artificial populations.

\section{Related work}
\paragraph{Learning and inheritance.}
The Baldwin effect concerns learning-mediated evolutionary change without directly encoding the acquired adjustment in offspring; direct writeback is a Lamarckian computational mechanism. Hinton and Nowlan provide a classic example of the former \cite{hinton1987}. Rivoire and Leibler explicitly compare ways of generating and transmitting variation, including phenotype-to-genotype feedback and environmental information \cite{rivoire2014}. Our use of separate inherited and lifetime tables instantiates a chosen architecture; it does not establish that architecture's biological existence or novelty.

\paragraph{Inherited learning in robots.}
Jelisavcic et al. compare inherited controllers with learning from scratch under different evaluation budgets \cite{jelisavcic2019}. Luo et al. investigate inherited learned traits, including newborn performance and a random-body comparison \cite{luo2023}. More recently, de Bruin et al. investigate parental sample reuse within Bayesian optimization under a low controller-learning budget \cite{debruinSamples}. These studies preclude interpreting reduced repeated learning as a new general finding. Our comparison instead preserves a candidate update's state-wise geometry, and our newborn intervention changes expressed action preferences while keeping inherited parameters fixed after the quota. The hereditary event quota here is also distinct from a controller's evaluation budget.

\paragraph{Learning and search allocation.}
Lamarckian and Baldwinian evolutionary algorithms can differ even on fixed objective functions. A recent empirical and runtime analysis identifies settings in which Baldwinian evolution outperforms Lamarckian evolution \cite{benito2026}. Thus an inheritance disadvantage in an externally stationary setting is not itself novel. Our event-allocation intervention differs from removing local learning: lifetime updates continue after new hereditary writes stop.

\paragraph{Ecology and benefit boundaries.}
Learning, resource competition, and reproduction already coexist in spatial agent models \cite{aguilar2019}. Thomsen and Rasmussen compare Darwinian, Baldwinian, and Lamarckian agents that acquire resources before reproducing, with outcomes affected by learning and replication costs \cite{thomsen2023}. Their fixed-size, nonspatial population differs from our variable occupancy and local resource depletion, but resource-dependent rankings are not new in themselves. Recent robot studies examine environmental conflict and predictability \cite{debruinDynamic}, and reduced inheritance benefits under pressure for morphological novelty \cite{muff2026}. Our external ecological rules remain fixed within a run; the boundary intervention is whether hereditary updating continues. We make no claim that these distinct experiments establish a common universal law. The novelty assessment is a targeted comparison with these works, not proof of the absence of other related designs.

\section{Model and experimental methods}
\subsection{Spatial ecology}
The lattice has $L=32$ sites per side and periodic boundaries. Each site contains at most one individual and a resource value $R_i\in[0,1]$. Initially, sites are independently occupied with probability $0.25$, occupied individuals have energy $E=1.5$, and all sites start with a resource value of one. Each sweep begins by replenishing every site's resource by $0.06$, capped at one, and then selects $L^2$ sites with replacement. Selecting an empty site does nothing. At an occupied site, background mortality occurs with probability $0.002$ before an action is chosen. Mortality is stochastic, with no fixed age limit; an individual may be selected multiple times or not at all in a sweep.

The available actions are rest, harvest, and divide, with respective energy costs $0.04$, $0.08$, and $0.12$. Harvest removes $\min(0.5,R_i)$ units of local resource and adds the same amount to the individual's energy, which is capped at four. A divide action succeeds if energy after its action cost exceeds two and at least one of the eight Moore neighbors is empty. An empty neighbor is selected uniformly. A fraction $f$ of the post-cost energy goes to the child and $1-f$ remains with the parent, conserving their total. Baseline $f=0.5$ is compared with $0.35$ or $0.65$ in specified experiments. Nonpositive post-action energy causes death. Individuals do not move.

\subsection{State, action selection, and lifetime learning}
Each individual carries an inherited table $g$ and a lifetime table $\theta$. Energy and resource are uniformly binned into eight bins each, and a binary variable records whether division would currently be feasible. The resulting $8\times8\times2=128$ states have three action values each. A sensitivity analysis uses $4\times4\times2$ states, with the same ecological engine. Both tables are initially zero in founders, and normally $\theta_{\rm child}=g_{\rm child}$ at birth.

With probability $0.05$ an individual chooses uniformly among the three actions; otherwise it chooses uniformly among actions with maximal values in the current row of $\theta$, treating values within $10^{-12}$ as ties. The selected coordinate is updated by
\begin{equation}
\theta(s,a)\leftarrow\theta(s,a)+\eta\{r-\theta(s,a)\},\qquad \eta=0.2.
\end{equation}
This is immediate action-value learning without a next-state bootstrap, not discounted-return Q-learning. Let $E_{\rm retained}$ be surviving post-action energy, including parent and child energy when division occurs, and zero after energy death. The reward is
\begin{equation}
r=E_{\rm retained}-E_{\rm before}+b\,\mathbf1_{\rm birth}-0.5\,\mathbf1_{\rm energy\ death},
\end{equation}
where the baseline birth bonus is $b=0.2$. Background deaths receive no action update. At a successful birth, the learning update precedes computation of the offspring's hereditary increment. Local reward is not population fitness, and a successful division can be locally reinforced even when repeated division has unfavorable ecological consequences.

\subsection{Hereditary update and random controls}
The original, or \emph{native}, writeback is
\begin{equation}\label{eq:native}
d=\lambda(\theta_{\rm parent}-g_{\rm parent}),\qquad
g_{\rm child}=g_{\rm parent}+d+\epsilon_{\mu}.
\end{equation}
Here $\epsilon_{\mu}$ denotes an independent mutation increment. All central experiments reported here set $\mu=0$, so $\epsilon_{\mu}=0$; they do not estimate a mutation--writeback phase diagram. Native writeback uses $\lambda\in\{0.25,0.5,1\}$. A no-writeback reference sets $\lambda=0$ while retaining lifetime learning. This reference is not an independent-mutation evolutionary algorithm.

For a nonzero candidate in the fixed-norm experiment, the transmitted step is rescaled to
\begin{equation}\label{eq:fixed}
d_q=q\frac{\theta-g}{\|\theta-g\|_2},\qquad q=0.1841261006694567.
\end{equation}
The value of $q$ was carried forward from the earlier study stage and held fixed across the reported confirmation, representation, and robustness conditions. It was not retuned for the coarser representation. In the fixed-norm implementation, candidate norms below $10^{-12}$ are treated as zero and skipped without consuming quota; no such skips occurred in the reported formal main conditions. Native updates have no corresponding minimum-norm floor. Equation~\ref{eq:fixed} is a direction intervention, not a reparameterization of a single fixed $\lambda$.

For any state's three candidate components $d_s$, decompose
\begin{equation}
m_s=\tfrac13\mathbf1^\top d_s,\qquad c_s=d_s-m_s\mathbf1.
\end{equation}
Our main random control samples $\phi_s$ uniformly from $[0,2\pi)$ in the two-dimensional zero-sum plane and constructs
\begin{equation}\label{eq:rotation}
\widetilde d_s=m_s\mathbf1+\|c_s\|_2(\cos\phi_s\,u+\sin\phi_s\,v),\quad
u=\frac{(1,-1,0)}{\sqrt2},\quad v=\frac{(1,1,-2)}{\sqrt6}.
\end{equation}
The operation preserves each row's mean, contrast norm, and full norm, and does not turn an all-zero row into a nonzero row. It randomizes action preference within states, not the states receiving updates. It does not preserve the sparsity of individual actions within a row.

This control retains state-level learning information. The matching is with the candidate from the same parent in that run; two evolving populations need not have equal state visitation, identical parents, or equal cumulative row-wise changes. We therefore use ``state-preserving randomization'' rather than an information-free mutation control. Whole-table shuffled and isotropic controls, each with the same total fixed norm, provide diagnostic comparisons. The shuffled control permutes candidate coordinates and applies independent random signs; the isotropic control normalizes an independent Gaussian vector. Near-zero candidates trigger an isotropic step in these weaker controls, rather than a skip; no near-zero candidate skips were recorded in the formal directed or state-preserving conditions. Their movement of updates across states is part of what the stronger control removes.

A quota $B$ limits the number of nonzero hereditary updates across the population, not births or learning interactions. After quota exhaustion, reproduction and lifetime learning continue, but offspring receive unchanged parental $g$. If a run uses all $B$ fixed-norm updates, its sum of update norms and sum of squared update norms are $Bq$ and $Bq^2$. These are not the net genome displacement and are not Shannon information. Early extinction can prevent quota exhaustion. Under native writeback even equal update counts do not equalize these magnitude totals.

\subsection{Newborn expression intervention}
The mechanism experiment uses fixed-norm updates and $B=512$. For each direction and seed, branches have identical histories until the quota is consumed. From the birth that consumes the 512th update onward, newborns either retain $\theta=g$ or have their expressed values replaced by the per-state mean,
\begin{equation}
\theta_{\rm child}(s,a)=\tfrac13\sum_{a'}g_{\rm child}(s,a').
\end{equation}
The latter erases newborn action preferences while preserving each state's mean value and the hereditary table itself. No further hereditary updates occur, and already-existing individuals are unchanged. For these newborns only, the learning rate is set to $\eta_c\in\{0,0.2,0.8\}$. This changes update weight without adding interactions or energy.

The main intervention contrast is erase minus keep at $\eta_c=0.2$ within each hereditary direction. Compensation is assessed by how the erase--keep loss changes at $\eta_c=0.8$. The zero-learning condition is an auxiliary limit. The two directions reach their quotas in different ecological states, so within-direction branch interventions do not constitute a complete mediation decomposition of the between-direction effect.

\subsection{Symmetric constraints and allocation diagnostics}
For the stability follow-up, both same-parent candidate updates receive a common global multiplier that caps their full norm at $Q=0.1841261006694567$. For bounded variants, each state's pair receives a further common multiplier that keeps both candidate offspring within the same coordinate range; a 1\% interior slack is used when a boundary constrains the step. We compare the original rule, the norm cap alone, the range $[-0.62,0.42]$, and the wider range $[-1,1]$. The narrower range follows the baseline reward bounds and zero initialization. These transformations preserve same-parent row matching but modify proportional writeback; they do not match the histories of different populations or their behavioral changes.

At $f=0.65$, $\lambda=1$, the common-time diagnostic retains the narrower range and compares continuous writeback, a 1,024-event quota, and common cutoffs after 150 or 750 complete sweeps. The times were fixed from earlier cutoff audits before the new seed block. Learning and inheritance of existing $g$ continue after new writes cease. The primary contrast is the change in the directed capped-time advantage under the 750-sweep rule relative to the event quota.

A separate allocation experiment normalizes each nonzero learning difference to $q=0.02$ before constructing its paired random candidate. This step size was fixed from the preceding update-size audit, not tuned on the new seeds. An opportunity is skipped without debiting the quota if either same-parent candidate cannot retain this norm inside the common narrower range. Random directions are drawn once per birth, not resampled until feasible. One schedule makes all 1,024 events available initially; the other releases 128 at internal times $0,107,214,321,429,536,643,750$, with unused credits carried forward until the run ends. The last tranche is available from sweep 751. The primary prediction is faster establishment under staged than earliest release within the directed condition; occupancy AUC is secondary. This is a fixed-step intervention, not a replication of the original proportional update magnitudes.

\subsection{Design, endpoints, and statistical unit}
All formal runs last 20,000 sweeps, with occupancy sampled every 200 sweeps including time zero. The primary endpoint is the first sampled time $t_j$ for which all five samples $t_j,\ldots,t_{j+4}$ have occupancy at least $0.95$. The five samples span 800 sweeps. Occupancy between samples is unobserved, and attainment does not guarantee that occupancy remains above the threshold thereafter. If no such span occurs, the endpoint remains unreached and the reported capped time $T^*$ is 20,000. This value is not an estimate of the unobserved time to eventual establishment.

We also report attainment counts, extinction, the trapezoidal area under the occupancy curve (AUC) divided by 20,000, and mean occupancy at the final five samples strictly after sweep 19,000. Thresholds 0.90 and 0.99 are auxiliary sensitivity endpoints. Runs are never removed for extinction, failure to establish, or incomplete quotas. Simulated descendants and repeated time points are not independent replicates.

The formal seed blocks are disjoint between study stages. Within a stage, the same seeds are paired across conditions, but shared seeds do not imply identical post-divergence ecological events. Comparisons use 20 seed pairs and 20,000 paired bootstrap resamples to obtain descriptive 95\% percentile intervals. Unless otherwise stated, bracketed ranges in the results denote these intervals. These intervals are not adjusted for multiple comparisons; we do not use them to claim family-wise significance, equivalence, or universal superiority.

\begin{table}[htbp]\centering\small
\caption{Formal experiment inventory. Counts are condition runs, not independent seeds. Shared no-writeback references are counted once per specified setting.}
\begin{tabular}{p{3.7cm}p{6.2cm}r}\toprule
Stage / seed block & Conditions & Runs\\\midrule
Control confirmation / 51000--51019 & Four update methods without quota, plus no writeback &100\\
Quota comparison / 52000--52019 & Four methods at five quotas; shared no writeback &420\\
Representation / 61000--61019 & Two binnings, two quotas, two directions; references &200\\
Expression / 62000--62019 & Two directions, two expressions, three newborn learning rates &240\\
Ecological sensitivity / 63000--63019 & Six settings, two directions and no writeback &360\\
Native connection / 81000--81019 & Two energy shares, two quota rules, three strengths, two directions; references &520\\
Symmetric stability / 101000--101019 & Four rules, six parameter settings, two directions; bounded quota follow-up &1200\\
Common cutoff / 102000--102019 & Four allocation rules, two directions &160\\
Fixed-budget release / 104000--104019 & Two release schedules, two directions &80\\\bottomrule
\end{tabular}\label{tab:inventory}
\end{table}

The synthesis contains 3,280 formal condition runs across nine blocks, including 1,440 runs in the three follow-ups. Development batches (45, 80, and 52 runs for the three archived experiment packages) and replay checks are excluded. The sequence was motivated by earlier exploratory results; it is not one prospectively registered study. Each later package records its local execution protocol and source hashes before its formal batch. The fixed-norm expression experiment, in particular, should not be represented as a mechanism confirmation performed under the native rule.

\section{Results}
We begin with the 512-event fixed-norm comparison, then test newborn expression in a separate seed block. Proportional-writeback and allocation experiments examine how far this interpretation extends; sensitivity analyses follow.

\subsection{Control geometry changes the apparent benefit}
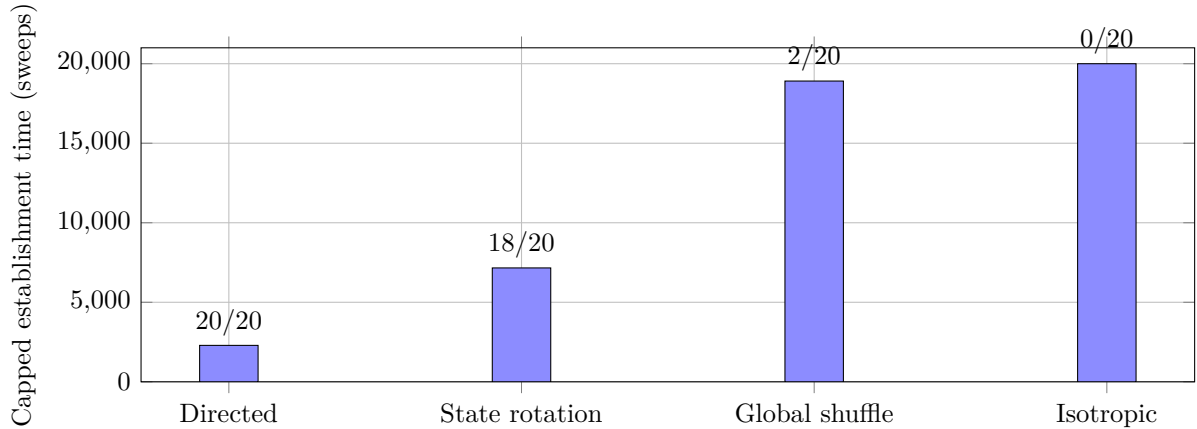
\begin{figure}[htbp]\centering
\begin{tikzpicture}\begin{axis}[width=0.94\linewidth,height=6cm,ymin=0,ymax=21000,scaled y ticks=false,ytick={0,5000,10000,15000,20000},ylabel={Capped establishment time (sweeps)},grid=major,legend style={at={(0.5,1.03)},anchor=south,legend columns=2,font=\small},tick label style={font=\small},label style={font=\small},ybar,bar width=22pt,xtick={1,2,3,4},xticklabels={Directed,State rotation,Global shuffle,Isotropic},nodes near coords,point meta=explicit symbolic,every node near coord/.append style={font=\small}]\addplot[fill=blue!45] coordinates {(1,2290) [20/20] (2,7160) [18/20] (3,18910) [2/20] (4,20000) [0/20]};\end{axis}\end{tikzpicture}
\caption{Fixed-norm comparison at $B=512$. Bars show mean capped time across 20 seeds; labels give the number establishing out of 20 runs. Failed runs remain at the 20,000-sweep cap, so these are not means conditional on success. The stronger random control sharply reduces the apparent attainment gap. All four groups use 512 updates with identical per-update norm.}\label{fig:control}
\end{figure}

At $B=512$, both directed inheritance and state-preserving randomization use their entire fixed-norm quotas. Directed inheritance establishes in 20/20 runs and the state-preserving random control in 18/20, with mean capped times of 2,290 and 7,160 sweeps, respectively. The paired random-minus-directed difference is 4,870 [3,040, 7,120] sweeps. Both methods have zero extinctions. In the same seed block, whole-table shuffling establishes in 2/20 runs and isotropic variation in 0/20 (Figure~\ref{fig:control}). Thus the magnitude of the apparent benefit depends strongly on the chosen random control. The 20/20 versus 18/20 outcome alone is not sufficient to assert a population-level difference in establishment probability.

The weak random controls allocate approximately one quarter of their squared update mass to physically unreachable represented states, whereas the main state-preserving control allocates none. This audit demonstrates a structural mismatch, but does not isolate unreachable mass as the sole reason for weaker performance: global shuffling and isotropic variation also change other aspects of the state--action structure.

\subsection{Newborn expression contributes to establishment}
At the standard newborn learning rate, retaining preferences in the directed branch yields 20/20 attainment and a mean time of 2,480 sweeps. Erasing preferences yields 6/20 attainment and a capped mean of 18,280 (Table~\ref{tab:mechanism}). Neither group becomes extinct. Erasure increases capped time by a paired mean of 15,800 [14,240, 17,070] sweeps and reduces normalized occupancy AUC by 0.23168 [0.20864, 0.25509]. This intervention concerns establishment and occupancy, not the necessity of preference expression for survival.

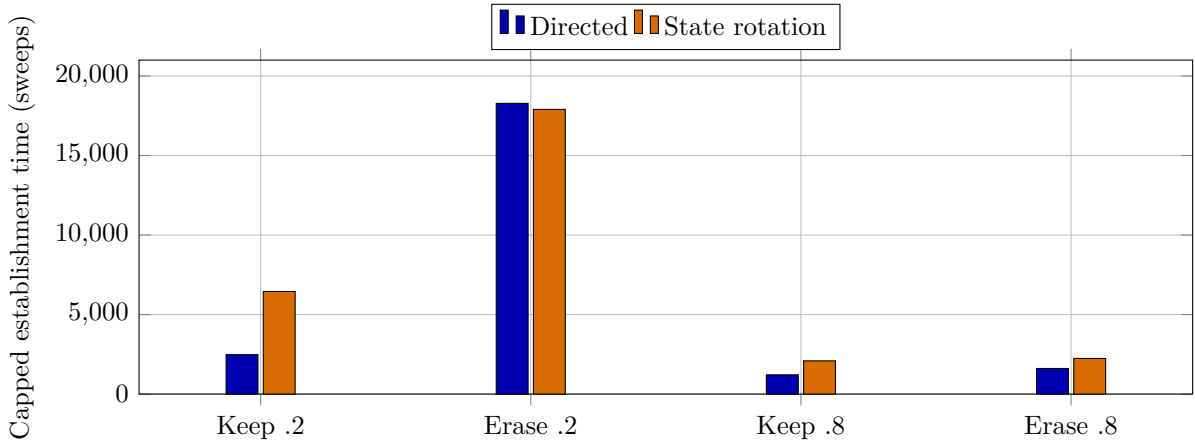
\begin{figure}[htbp]\centering
\begin{tikzpicture}\begin{axis}[width=0.94\linewidth,height=6cm,ymin=0,ymax=21000,scaled y ticks=false,ytick={0,5000,10000,15000,20000},ylabel={Capped establishment time (sweeps)},grid=major,legend style={at={(0.5,1.03)},anchor=south,legend columns=2,font=\small},tick label style={font=\small},label style={font=\small},ybar,bar width=12pt,xtick={1,2,3,4},xticklabels={Keep .2,Erase .2,Keep .8,Erase .8},enlarge x limits=0.15]\addplot[fill=blue!70!black] coordinates {(1,2480) (2,18280) (3,1210) (4,1610)};\addplot[fill=orange!85!black] coordinates {(1,6450) (2,17900) (3,2090) (4,2240)};\legend{Directed,State rotation}\end{axis}\end{tikzpicture}
\caption{Expression and learning-rate intervention beginning with the birth that consumes the 512th fixed-norm update. The labels specify newborn expression and $\eta_c$. Each bar includes all 20 runs; attainment counts and auxiliary zero-learning results are in Table~\ref{tab:mechanism}. The same-direction branches share their pre-intervention histories.}\label{fig:mechanism}
\end{figure}

Increasing the newborn learning rate to $0.8$ restores attainment to 20/20 in the erased directed branch, with mean time 1,610. The keep branch at that rate has a mean establishment time of 1,210 sweeps. The erase--keep capped-time loss consequently falls from 15,800 to 400 sweeps, a difference-in-differences of 15,400 [13,860, 16,680]. The random branch also benefits from faster relearning (Figure~\ref{fig:mechanism}). These results support a contribution from the burden of reconstructing newborn action preferences. They do not establish that the entire directed advantage is mediated by this burden: the between-direction AUC interaction interval includes zero, and the native rule is not tested by this intervention.

At $\eta_c=0$, erased populations all become extinct. Keeping preferences yields 20/20 attainment for directed inheritance and 6/20 for the random control. Moreover, the directed keep group is faster at rate zero than at $0.2$. Learning-rate effects therefore cannot be summarized as monotonic improvements from more learning.

\begin{table}[htbp]\caption{All expression-intervention conditions. D denotes directed inheritance and R denotes state-preserving randomization. Reached entries give successful runs out of 20; AUC D/R lists the two normalized occupancy areas. Times are capped at 20,000 sweeps.}\label{tab:mechanism}
\begin{center}
\small
\begin{tabular}{rlccrrc}
\toprule
$\eta_c$ & Expression & Reached D & Reached R & $\bar T_D^*$ & $\bar T_R^*$ & AUC D/R \\
\midrule
0 & Keep & 20/20 & 6/20 & 1390 & 15540 & 0.959/0.379 \\
0 & Erase & 0/20 & 0/20 & 20000 & 20000 & 0.009/0.009 \\
0.2 & Keep & 20/20 & 18/20 & 2480 & 6450 & 0.934/0.827 \\
0.2 & Erase & 6/20 & 5/20 & 18280 & 17900 & 0.702/0.649 \\
0.8 & Keep & 20/20 & 20/20 & 1210 & 2090 & 0.966/0.934 \\
0.8 & Erase & 20/20 & 20/20 & 1610 & 2240 & 0.955/0.929 \\
\bottomrule
\end{tabular}
\end{center}
\end{table}

\subsection{Continuous native writeback retains a benefit}
We next remove fixed-norm rescaling and use Equation~\ref{eq:native}. With continuous writeback, both directions establish in all 20 runs for each of the six tested $f\times\lambda$ combinations. Directed inheritance is faster in each of the 20 paired seeds within every combination, and also yields higher normalized occupancy AUC. All native settings and their occupancy outcomes are reported in Table~\ref{tab:native}.

The benefit therefore does not occur only under fixed-norm rescaling. This native comparison does not, however, isolate direction at equal cumulative variation: different histories produce different update magnitudes and counts. The no-writeback references become extinct in all 20 seeds for each of these two environments; this is a conditional reference outcome, not a claim that hereditary learning is always necessary for establishment.

\subsection{Stopping writeback can change the ranking}
At $f=0.65$, continuous native writeback favors directed inheritance for all three strengths, but stopping at $B=1024$ changes the outcome. At $\lambda=0.5$, directed/random attainment is 0/20 versus 17/20; at $\lambda=1$, it is 5/20 versus 20/20. At $\lambda=0.25$, neither group establishes, yet AUC is lower for directed inheritance (0.686 versus 0.734). None of these main variation groups becomes extinct. Quota-limited and continuous branches have identical update records through the 1,024th update for all 240 matched pairs, validating the common prefix of this intervention.

There is also an endpoint-specific reversal at baseline $f=0.5$, $\lambda=0.25$, and $B=1024$: directed/random attainment is 7/20 versus 20/20, and capped times are 17,790/4,900. Yet directed AUC is slightly higher (0.844 versus 0.830), while its late occupancy is lower (0.948 versus 0.988). At the auxiliary 90\% threshold both establish in all seeds; at 99\%, attainment is 0/20 versus 8/20. These observations do not support a statement that random updating is better on every metric.

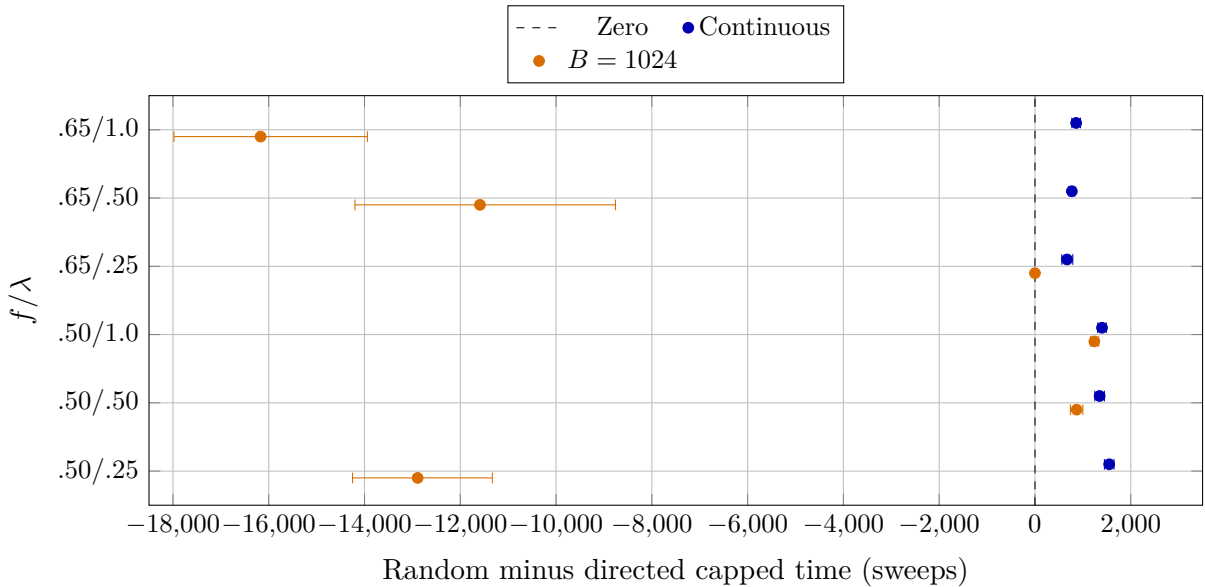
\begin{figure}[htbp]\centering
\begin{tikzpicture}\begin{axis}[width=.94\linewidth,height=7cm,xmin=-18500,xmax=3500,scaled x ticks=false,xlabel={Random minus directed capped time (sweeps)},ytick={1,2,3,4,5,6},yticklabels={$.50/.25$,$.50/.50$,$.50/1.0$,$.65/.25$,$.65/.50$,$.65/1.0$},ylabel={$f/\lambda$},ymin=.5,ymax=6.5,grid=major,legend style={at={(0.5,1.03)},anchor=south,legend columns=2,font=\small},tick label style={font=\small}]\addplot[black,dashed] coordinates {(0,.5)(0,6.5)};\addplot[only marks,mark=*,color=blue!70!black,error bars/.cd,x dir=both,x explicit] coordinates {(1550,1.1) += (100,0) -= (100,0) (1350,2.1) += (100,0) -= (100,0) (1400,3.1) += (90,0) -= (90,0) (670,4.1) += (120,0) -= (110,0) (770,5.1) += (60,0) -= (60,0) (860,6.1) += (90,0) -= (90,0)};\addplot[only marks,mark=*,color=orange!85!black,error bars/.cd,x dir=both,x explicit] coordinates {(-12890,0.9) += (1560,0) -= (1360,0) (870,1.9) += (130,0) -= (130,0) (1240,2.9) += (90,0) -= (80,0) (0,3.9) += (0,0) -= (0,0) (-11590,4.9) += (2830,0) -= (2610,0) (-16170,5.9) += (2230,0) -= (1810,0)};\legend{Zero,Continuous,$B=1024$}\end{axis}\end{tikzpicture}
\caption{Native writeback: paired differences in capped establishment time. Positive values favor directed inheritance. Points are means across 20 seed pairs; intervals are descriptive 95\% paired bootstrap intervals, without multiplicity correction. The zero difference for $f=0.65$, $\lambda=0.25$, $B=1024$ reflects complete nonattainment in both groups, not equivalence.}\label{fig:native}
\end{figure}

\begin{table}[htbp]\caption{All native variation conditions (20 seeds per direction). Each paired entry lists the directed value followed by the random value. In the Reached column, these are counts out of 20 in each group; for example, 20/20 means that both groups have 20 successful runs. $B=\infty$ denotes continuous writeback. All listed groups have zero extinctions. Late is the final-five-sample occupancy mean.}\label{tab:native}
\begin{center}
\small
\begin{tabular}{rrrcccc}
\toprule
$f$ & $B$ & $\lambda$ & Reached D/R & $\bar T^*$ D/R & AUC D/R & Late D/R \\
\midrule
0.5 & $\infty$ & 0.25 & 20/20 & 2490/4040 & 0.927/0.856 & 0.999/0.999 \\
0.5 & $\infty$ & 0.5 & 20/20 & 1490/2840 & 0.957/0.908 & 0.999/0.999 \\
0.5 & $\infty$ & 1.0 & 20/20 & 930/2330 & 0.975/0.927 & 0.999/0.999 \\
0.5 & 1024 & 0.25 & 7/20 & 17790/4900 & 0.844/0.830 & 0.948/0.988 \\
0.5 & 1024 & 0.5 & 20/20 & 2040/2910 & 0.939/0.906 & 0.990/0.998 \\
0.5 & 1024 & 1.0 & 20/20 & 1000/2240 & 0.973/0.930 & 0.998/0.999 \\
0.65 & $\infty$ & 0.25 & 20/20 & 2060/2730 & 0.970/0.917 & 0.999/0.999 \\
0.65 & $\infty$ & 0.5 & 20/20 & 1170/1940 & 0.980/0.943 & 0.999/0.999 \\
0.65 & $\infty$ & 1.0 & 20/20 & 660/1520 & 0.986/0.958 & 0.999/0.999 \\
0.65 & 1024 & 0.25 & 0/0 & 20000/20000 & 0.686/0.734 & 0.718/0.819 \\
0.65 & 1024 & 0.5 & 0/17 & 20000/8410 & 0.736/0.897 & 0.778/0.967 \\
0.65 & 1024 & 1.0 & 5/20 & 17720/1550 & 0.870/0.956 & 0.904/0.997 \\
\bottomrule
\end{tabular}
\end{center}
\end{table}

\subsection{Benefits survive common stability constraints}
The new stability seed block retains the continuous directed speed advantage in all six strength-by-energy settings under the narrower common range. Both directions establish in all runs, with no extinctions; directed inheritance is faster in every seed pair. The mean directed/random times at $f=0.5$ are 2,500/4,040, 1,610/3,040, and 1,230/3,040 for $\lambda=0.25,0.5,1$, respectively; at $f=0.65$ they are 2,030/2,810, 1,180/1,990, and 1,000/1,890. Directed AUC is also higher in each condition. The wider range and norm-only variants retain the same time ordering across their tested continuous conditions.

The unbounded random rule in this new block reaches a maximum single step of 65.381 and a maximum absolute offspring coordinate of 128.954. The norm cap limits the former but still permits coordinates up to 3.426; the narrower range limits absolute coordinates to 0.620. Thus continuous benefits do not require unbounded random drift. Nevertheless, at $f=0.65$, $\lambda=1$, bounded continuous directed/random cumulative norm sums are approximately 3,593/7,492. Matching candidate geometry within a parent does not equalize realized population-wide variation.

\subsection{The event cutoff is endogenous to reproduction}
The bounded 1,024-event comparison still reverses at $f=0.65$, $\lambda=1$: directed attainment is 0/20 and random attainment 14/20 in the stability block. Its post hoc audit finds mean quota-exhaustion times of 133.95 and 724.00 sweeps. Thus the two methods receive different temporal opportunities despite the same event quota. Stopping new writes neither deletes accumulated hereditary values nor stops lifetime learning.

The subsequent independent cutoff block repeats this dependence (Table~\ref{tab:cutoff}). The prespecified 750-sweep contrast increases the random-minus-directed capped-time difference by 15,930 sweeps [13,470,18,190] relative to the 1,024-event rule. Within the 750-sweep condition, the directed advantage is 6,520 [4,400,8,990]. However, this change accompanies approximately 3,910 directed writes versus 1,094 random writes, so it is not a pure timing effect at equal update amounts. The early common cutoff leaves both groups unestablished, although directed AUC is higher. All four conditions have zero extinctions.

\begin{table}[htbp]\centering\small
\caption{Common-cutoff diagnostic at $f=0.65$, $\lambda=1$; 20 runs per direction. Entries list directed/random values; attainment entries are two counts, each out of 20. Failed runs retain the 20,000-sweep time cap.}\label{tab:cutoff}
\begin{tabular}{lrrrr}\toprule
Rule & Attained D/R & $\bar T^*$ D/R & AUC D/R & Mean writes D/R\\\midrule
Continuous & 20/20 & 1000/1870 & .982/.947 & 46888/47796\\
1,024 events & 0/14 & 20000/10590 & .800/.898 & 1024/1024\\
150 sweeps & 0/0 & 20000/20000 & .814/.688 & 1073/444\\
750 sweeps & 20/18 & 1000/7520 & .979/.907 & 3910/1094\\\bottomrule
\end{tabular}
\end{table}

\subsection{Matched-budget staging improves occupancy, not establishment}
All 80 fixed-step allocation runs use their full quota, giving identical cumulative norm sums of 20.48 and squared-norm sums of 0.4096. No run becomes extinct, but none reaches the primary establishment threshold. Consequently, the prespecified faster-establishment prediction is not supported, and the capped-time tie is not evidence of equivalence.

Secondary occupancy outcomes do differ (Table~\ref{tab:staging}). Staging improves directed AUC by 0.08249 [0.07842,0.08664] and random AUC by 0.00376 [$-0.02953$,0.03276]. The difference between these improvements is 0.07873 [0.05031,0.11112]. These descriptive results cannot be attributed to extra cumulative update magnitude, but they still combine timing, state coverage, geometric feasibility, and selection along diverging histories. They support a limited allocation effect rather than a successful establishment intervention.

\begin{table}[htbp]\centering\small
\caption{Fixed-step allocation ($q=0.02$, $B=1024$). Each group has 20 seeds, zero attainment and zero extinction. All groups use exactly the same event count and cumulative magnitudes.}\label{tab:staging}
\begin{tabular}{llrrr}\toprule
Schedule & Direction & AUC & Late occupancy & Mean exhaustion sweep\\\midrule
Earliest & Directed & .6786 & .7010 & 142.3\\
Earliest & Random & .6274 & .7140 & 803.7\\
Staged & Directed & .7611 & .7919 & 777.9\\
Staged & Random & .6312 & .7339 & 1416.2\\\bottomrule
\end{tabular}
\end{table}

\subsection{Quota and representation sensitivity}
In the fixed-norm confirmation without an update quota, directed inheritance and state-preserving randomization both establish in 20/20 runs. Their mean establishment times are 1,290 and 3,040 sweeps, respectively. The paired mean random-minus-directed difference is 1,750 [1,650, 1,850] sweeps. Because update counts differ in this continuous experiment, this is not a comparison at equal cumulative variation.

Table~\ref{tab:fixed} shows all main fixed-norm quota conditions. At $B=128$, both directions become extinct in all runs. At $B=256$, neither establishes, although 17 directed and seven random populations survive. Equal capped times therefore conceal different outcomes. At $B=1024$ and $2048$, both establish in every run, with directed mean times of 1,440 and 1,300 compared with 3,080 and 3,010. These results support a speed benefit over part of the tested quota range, not a universal survival advantage. Both methods first exceed 80\% sample attainment at $B=512$ on this grid, so the data do not demonstrate a halving of the required quota.

\begin{table}[htbp]\caption{Fixed-norm quota comparison. D: directed; R: state-preserving random. Each group has 20 seeds. Times are sweeps; unreached runs contribute 20,000. Extinct D/R lists the extinction count in each group, not a fraction.}\label{tab:fixed}
\begin{center}
\small
\begin{tabular}{rccrrc}
\toprule
$B$ & Reached D & Reached R & $\bar T_D^*$ & $\bar T_R^*$ & Extinct D/R \\
\midrule
128 & 0/20 & 0/20 & 20000 & 20000 & 20/20 \\
256 & 0/20 & 0/20 & 20000 & 20000 & 3/13 \\
512 & 20/20 & 18/20 & 2290 & 7160 & 0/0 \\
1024 & 20/20 & 20/20 & 1440 & 3080 & 0/0 \\
2048 & 20/20 & 20/20 & 1300 & 3010 & 0/0 \\
\bottomrule
\end{tabular}
\end{center}
\end{table}

Changing to the coarser $4\times4\times2$ representation retains the time ordering in the independent representation block. At $B=512$, the directed/random means are 1,180/1,960 sweeps; at $B=1024$, they are 1,020/1,600, with 20/20 attainment in all four groups. This is a binning-sensitivity result using the same simulator and learner family, not independent ecosystem or algorithmic replication.

\subsection{Ecological and numerical limits}
Fixed-norm sensitivity experiments at $B=1024$ preserve the speed ordering at lower mortality, smaller offspring energy share, and zero birth bonus (Appendix~\ref{app:robust}). At background mortality $0.01$, both directions become extinct before using the quota: average update counts are 275.8 and 225.2. At low mortality $0.0005$, even the no-writeback reference establishes in 20/20 runs. The benefit is therefore neither necessary in every environment nor sufficient under severe mortality.

In the fixed-norm high-offspring-share condition, directed/random attainment is 0/20 versus 18/20. Both survive, but directed occupancy is lower. The native continuous results show why this cannot be attributed to offspring energy share alone: the reversal depends on the update rule and cutoff as well. Higher birth and energy-death counts in the fixed-norm directed group are post hoc descriptions; their causal role has not been independently established.

The randomized counterpart of the original proportional rule lacks the directed rule's convex-hull property and can produce much larger steps. The bounded follow-up rules out unbounded random drift as necessary for a continuous benefit, but does not equalize cumulative variation across populations. Full magnitude and numerical-replay diagnostics are retained in Appendix~\ref{app:nativeaudit}.

\section{Discussion}
The main result is a conditional establishment-speed benefit whose interpretation changes with the control. Equal Euclidean norm does not preserve where an update acts: whole-table randomization mixes action direction with state allocation and reachability. Preserving state-wise geometry narrows that mismatch, but retains some learning information and does not match changes in action probabilities. Extending this control to other representations requires preserving the structure relevant to their behavior.

Newborn erasure and learning-rate compensation support a contribution from reduced relearning in the fixed-norm regime. The branches preserve inherited tables and row means, and differ from the quota-exhausting birth onward. They therefore test preference expression after a shared within-direction history. These interventions neither remove all inherited information nor establish complete mediation of the between-direction advantage; they also do not identify the mechanism of native proportional writeback.

Allocation experiments show why equal event budgets require temporal audits. Faster reproduction can exhaust a quota earlier; common-time cutoffs change this coupling but also change realized update amounts. Matching counts and cumulative magnitudes supports an allocation-policy effect on occupancy, although timing, state coverage, geometric feasibility, and selection remain coupled. The failed establishment prediction is retained. Moreover, the quota is an experimental constraint, not a modeled metabolic cost of inheritance.

Transfer is limited by the simple ecology, immediate-value learner, engineered reward, explicit division-feasibility input, and absence of independent mutation in the core comparisons. Mortality is stochastic rather than age-based. Coarser binning uses the same simulator, so it is not independent replication. Establishment is observed only at sampled times; auxiliary thresholds do not resolve between-sample fluctuations. Step sizes were fixed after exploratory development rather than systematically varied here. Each formal condition has 20 seeds, and descriptive intervals are not corrected for multiple comparisons. The full program was not prospectively registered, although later batches had frozen local protocols. Independent implementation and prespecified sensitivity tests are the next checks on generality; the current evidence does not establish increased long-term evolvability.

\section{Conclusion}
Inherited action preferences can help a population retain useful experience as individuals die and are replaced, but their population-level value depends on how that experience is expressed and transmitted. In our resource-limited artificial ecology, state-preserving random controls reduce a large apparent establishment advantage to a conditional speed benefit. Newborn preference erasure and compensation by faster learning support a contribution from reduced relearning, while continuous proportional inheritance retains a speed advantage under common stability constraints.

The allocation of hereditary updates also shapes the measured benefit. Event quotas and shared time cutoffs can produce different rankings while changing realized update amounts. When update count and cumulative magnitudes are matched, staged release improves time-averaged occupancy, but no group reaches the prespecified establishment threshold. Our contribution is a framework of controls and interventions for evaluating inherited learning: comparisons must account for update structure and stability, newborn expression, and the coupling between reproduction and update allocation. The evidence supports conditional benefits in this artificial ecology. Testing whether these distinctions predict outcomes in an independently implemented learning system is the next step toward broader applicability.

\section*{Data and code availability}
Code, protocols, formal seed-level results, population traces, summaries, and validation records are available at \url{https://github.com/deep-geo/inherited-learning-ecology}. Full per-event logs are retained separately and are available on request via the email address listed on the title page. No permanent archive DOI has yet been assigned.

\section*{Author declarations}
This research received no funding. We declare no competing interests. The experiments are computational simulations, with no human participants, animals, or biological cell experiments. We take responsibility for the scientific content.

\appendix
\section{Additional ecological outcomes}\label{app:robust}
Table~\ref{tab:robust} reports the six fixed-norm sensitivity settings, each at $B=1024$. The baseline mortality is $0.002$, offspring share $0.5$, and birth bonus $0.2$. Each alternative changes one of these settings: mortality to $0.0005$ or $0.01$, offspring share to $0.35$ or $0.65$, or bonus to zero. Changing offspring share also changes the complementary parent share. No-writeback results and individual runs are retained in the accompanying data, not pooled across settings. All high-mortality failures are retained despite unused quota.
\begin{table}[htbp]\caption{Ecological sensitivity. Paired entries list directed and random values; each group has 20 seeds. Reached and Extinct entries are two counts, each out of 20, not fractions. The high-mortality capped-time tie reflects nonattainment in both groups, not equivalence.}\label{tab:robust}
\begin{center}
\small
\begin{tabular}{lcccc}
\toprule
Setting & Reached D/R & $\bar T^*$ D/R & Extinct D/R & AUC D/R \\
\midrule
baseline & 20/20 & 1460/3190 & 0/0 & 0.960/0.895 \\
low death & 20/20 & 1000/1730 & 0/0 & 0.978/0.959 \\
high death & 0/0 & 20000/20000 & 20/20 & 0.002/0.002 \\
low child energy & 20/20 & 1530/3160 & 0/0 & 0.953/0.903 \\
high child energy & 0/18 & 20000/5290 & 0/0 & 0.814/0.908 \\
no birth bonus & 20/20 & 2160/3560 & 0/0 & 0.931/0.881 \\
\bottomrule
\end{tabular}
\end{center}
\end{table}

\section{Native magnitude and numerical audit}\label{app:nativeaudit}
Native hereditary magnitude is not matched over trajectories. For $f=0.5$, $\lambda=0.5$, and $B=1024$, the directed/random mean sums of increment norms are approximately 99.10/158.09 and the mean sums of squared norms are 13.28/30.03. Under continuous writeback at $f=0.65$, $\lambda=1$, the random squared-sum median is approximately 15,343 versus 567 for directed inheritance. The random group's mean squared increment sum, approximately 480,899, is dominated by a run whose squared sum is $9.31\times10^6$ and maximum single-step norm is approximately 1,664.48. This run is retained.

For $0\leq\lambda\leq1$, directed native inheritance is the coordinate-wise convex combination $(1-\lambda)g+\lambda\theta$. State rotation preserves candidate row geometry but not this convex-hull property. Differential stability and variation magnitude therefore limit interpretation of the original native comparison. The bounded follow-up rules out unbounded random drift as necessary for a continuous benefit, but does not decompose the original effect or equalize cumulative changes. A diagnostic replay of the large-step run passes row-scaled floating-point tolerances and reproduces its saved outputs exactly; numerical validity does not eliminate this scientific limitation.

\section{Reproducibility and provenance}
Codex, a generative AI tool, assisted with simulation code and execution, analysis, literature discovery, figures, and manuscript preparation. We are responsible for the scientific content. The numerical and implementation checks described below document the validation performed; AI assistance is not independent verification.

Simulation code is written in C++17 with separate pseudorandom streams for ecological events, independent noise, direction generation, and measurements. Seeds and source snapshots are included per stage. Matched seeds align initial conditions and random-stream initialization, but do not enforce identical encounters after trajectories diverge. Source and protocol hashes were recorded by the experiment runners. Existing checks include population accounting, resource bounds, energy split conservation, geometry preservation, zero-learning and zero-writeback comparisons, and exact replay of representative outputs.

The zero-learning check compares writeback variants within the same learning-rate setting, not zero learning against learning without writeback. An initially incorrect implementation-check comparison was corrected before the native formal batch; the failed diagnostic is retained in its validation records. The large native random-update replay required a scale-relative numerical tolerance rather than the original absolute $10^{-10}$ row-norm threshold. Its maximum reported absolute discrepancy was $4.66\times10^{-10}$; a diagnostic run with row-scaled tolerances reproduced all seven archived output types byte for byte.

For the initial manuscript synthesis, all 1,840 formal population traces were independently re-read to recompute attainment at 90\%, 95\%, and 99\%, capped times, normalized occupancy AUC, late occupancy, and extinction. These values agree with the archived seed summaries. This is a data-to-manuscript consistency check, not an independent reimplementation or a new simulation experiment. The original paired bootstrap intervals are retained. Development and replay rows are excluded from formal effect estimates. The 1,440 subsequent formal runs underwent corresponding population, energy, update-ledger, geometry, and bound checks. The cutoff implementation was checked against the preceding simulator with its new options disabled, and fixed-step release checks verified all realized quotas and magnitudes. The new seed blocks were not extended after outcome inspection.

\section{Scope of the literature comparison}
The related-work assessment uses targeted searches and primary texts available through 30 September 2026. It is not a systematic review and does not substantiate a first-ever claim. The 2023 robot preprint and its journal publication are treated as one work. The Bayesian-optimization sample-inheritance work is cited by its verified 2026 arXiv version, although an author data record exists from 2025; the upload date is not asserted to be its earliest dissemination. The two other 2026 robot works are cited as preprints without asserting final conference acceptance. Closely related effects and mechanisms are acknowledged rather than recast as newly discovered principles.

\end{document}